\documentclass[letterpaper, 10 pt, conference]{ieeeconf}

\IEEEoverridecommandlockouts    

\usepackage{cite}
\usepackage{bm}

\usepackage{xcolor}

\makeatletter

\let\labelindent\relax
\usepackage{multicol}
\usepackage{multirow}
\usepackage{amsthm}

\usepackage{amsmath}
\usepackage{amsfonts}
\usepackage{amssymb}
\usepackage[export]{adjustbox}

\usepackage{tikz}
\usetikzlibrary{positioning}
\usepackage{xcolor}
\definecolor{processblue}{cmyk}{0.96,0,0,0}
\usepackage{caption}
\usepackage{subcaption}
\usepackage{enumitem}

\theoremstyle{definition}
\newtheorem{theorem}{Theorem}[section]

\newtheorem{lemma}[theorem]{Lemma}

\newtheorem{definition}[theorem]{Definition}

\newtheorem{remark}{Remark}[section]
\theoremstyle{remark}

\usepackage{graphicx}
\usepackage{accents}

\usepackage{algorithm, algpseudocode, xspace}
\usepackage{mathtools}

\usepackage{algpseudocode}
\algrenewcommand\textproc{}
\usepackage[english]{babel}
\usepackage[utf8]{inputenc}
\algnewcommand{\parState}[1]{\State%
    \parbox[t]{\dimexpr\linewidth-\algmargin}{\strut\hangindent= \algorithmicindent \hangafter=1 #1\strut}}

\usepackage{array}
\usepackage{pdfpages}
\usepackage{tikz}
\usetikzlibrary{positioning}
\usepackage{mathtools}
\allowdisplaybreaks

\begin{document}
\title{\LARGE \bf Bayesian meta learning  for trustworthy uncertainty quantification}

\author{Zhenyuan Yuan$^{1}$ \qquad Thinh T. Doan$^{1}$
	\thanks{$^{1}$Zhenyuan Yuan and Thinh T. Doan are with Bradley Department of Electrical and Computer Science, Virginia Tech, Blacksburg, VA 24061, USA (email:
		{\tt\small \{zyuan18,thinhdoan\}@vt.edu}). This work was supported by the National Science Foundation under ECCS-CAREER Grant No. 2339509.}%
}


\maketitle
\thispagestyle{plain}
\pagestyle{plain}
\IEEEpeerreviewmaketitle


\begin{abstract} 
	We consider the problem of Bayesian regression with trustworthy uncertainty quantification.  We define that the uncertainty quantification is trustworthy if the ground truth can be captured by intervals dependent on the predictive distributions with a pre-specified probability. Furthermore, we propose, \textsf{Trust-Bayes},  a novel optimization framework for  Bayesian meta learning  which is cognizant of trustworthy uncertainty quantification without explicit assumptions on the prior model/distribution of the functions.   We characterize the lower bounds of the probabilities of the ground truth being captured by the specified intervals and analyze the sample complexity with respect to the feasible probability for trustworthy uncertainty quantification.  Monte Carlo simulation  of a case study using Gaussian process regression is conducted for verification and comparison with the \textsf{Meta-prior} algorithm.
\end{abstract}


\IEEEpeerreviewmaketitle

\section{Introduction}
Engineering systems operating in the real world are usually subject to unknown uncertainties. Examples include autonomous cars driving in urban scenarios, unmanned aerial robots for outdoor package delivery and mobile robots for search-and-rescue. In order to ensure mission success and the safety of these systems while maintaining high autonomy, it is necessary to learn and quantify these uncertainties in a trustworthy manner such that these systems can operate with minimal   human supervision/intervention.

Bayesian learning \cite{mitchell1997machine} is a class of statistical learning frameworks, including but not limited to Gaussian process regression (GPR) \cite{williams2006gaussian}, Kalman filtering \cite{welch1995introduction}, Bayesian neural network \cite{lampinen2001bayesian} and particle filtering \cite{bishop2006pattern}. In general, Bayesian learning first models a target (e.g., system state, parameter or function) as a sample from a distribution a priori, then given a set of data it utilizes the Bayesian inference framework to compute a posterior distribution of the target for prediction. 
With proper choice of the prior distribution and mild assumptions on the target, Bayesian learning is able to consistently approximate the target \cite{choi2007posterior}\cite{box2011bayesian}. Established analysis, such as the well-known PAC-Bayesian theorems \cite{mcallester1999some}, has shown that the generalization error for the performances of Bayesian learning methods decreases at the rate of $\mathcal{O}(\frac{1}{\sqrt{n}})$, where $n$ is the number of data samples. Furthermore, the predictive distributions, i.e., the prior and the posterior distributions, inherently allow Bayesian models to  predict with uncertainty quantification for each input. These aforementioned advantages make Bayesian learning a powerful tool in a variety of applications, e.g., optimization \cite{garnett2023bayesian}, learning-based control \cite{lederer2022cooperative,liu2018gaussian,virani2018imitation}, motion planning\cite{mukadam2016gaussian}\cite{yuan2022dslap}, state estimation \cite{welch1995introduction}  and system identification \cite{chiuso2012bayesian}.

Based on the Bayes rule \cite{mitchell1997machine}, the posterior distribution reflects predictive uncertainties accurately only if the likelihood function and the prior distribution are correctly specified, which are also a common assumption in the analysis for uncertainty quantification \cite{srinivas2012information}\cite{fisac2019general}\cite{yuan2024lightweight}. However, in some cases, the aforementioned information may not be obtained a priori accurately. 
To relax this assumption,  existing works instead assume the likelihood and/or the prior distribution are specified structurally such that the hyperparameters can be learned through data \cite{williams2006gaussian}\cite{wilson2016deep}\cite{shen2019random}. While there are powerful function approximation models, such as deep neural networks and finite-dimensional basis functions, which are able to consistently approximate a wide range of possible models, in some cases it can be hard to ensure the selected class of approximation models fully capture the target when there is not much prior information about the structure of the target. Furthermore, when the data is scarce, the hyperparameters may not be well learned \cite{vapnik2013nature}.
As a result, the uncertainty quantification obtained from the predictive distributions may not be trustworthy, and the subsequent operations, e.g., synthesis of safety controller \cite{zhou2020general}\cite{umlauft2020smart}  and selection of safe decisions \cite{berkenkamp2023bayesian}\cite{sui2018stagewise}, leveraging the uncertainty quantification can be unreliable. 

Meta learning is a machine learning framework which aims to utilize the data of a collection of tasks to identify a good initialization/prior for the learning algorithm in new tasks such that fast learning can be achieved using a small amount of data \cite{finn2017model}. The process of identifying a good initialization/prior is known as the {\em meta training} procedure, and the learning in a new task is known as the {\em adaptation} procedure. The problem is usually formulated as an
optimization problem, where the objective function is
the expected performance of the adapted model in a
new task. Most of the works do not consider uncertainty quantification \cite{finn2017model,rajeswaran2019meta,khodak2019adaptive,lee2021meta}. Generalization errors, which can be used to derive for uniform uncertainty quantification, are considered in \cite{xu2024online}\cite{fallah2021generalization} for adapted models. Bayesian meta learning is considered in \cite{yoon2018bayesian,zhang2021shallow, lew2022safe,harrison2020meta,rothfuss2021pacoh,patacchiola2020bayesian,nabi2022bayesian}, where the posterior distributions provide input-dependent uncertainty quantification. In particular, papers \cite{yoon2018bayesian}\cite{zhang2021shallow} consider meta learning for Bayesian neural networks, where the prior distributions of the parameters in the neural networks are meta trained to optimize the performances of the neural networks with parameters sampled from the posterior distributions. Papers
\cite{lew2022safe,harrison2020meta, rothfuss2021pacoh, patacchiola2020bayesian, nabi2022bayesian} consider Gaussian process or Bayesian linear regression, where the hyperparameters in the prior covariance and/or mean (functions) are meta trained provided pre-specified structures. Prediction accuracy is considered in these methods, however, whether the learned prior and posterior provide trustworthy uncertainty quantification remains an open question.

{\bf Contribution statement.} In this paper, we consider the problem of Bayesian regression with trustworthy uncertainty quantification.  We propose a Bayesian meta learning framework which is cognizant of trustworthy uncertainty quantification without explicit prior assumptions on the model/distribution of the functions. Specifically, we define that the uncertainty quantification is trustworthy if the ground truth can be captured by intervals dependent on the predictive distributions with a pre-specified probability. We then propose \textsf{Trust-Bayes}, a novel optimization framework for Bayesian meta learning with constraints on trustworthy uncertainty quantification using the meta-trained prior distribution and the posterior distribution. We characterize the lower bounds of the probabilities of the ground truth being captured by the specified intervals in terms of the empirical estimates from meta training. We further analyze the sample complexity with respect to the feasible pre-specified probability  for trustworthy uncertainty quantification. In summary, our major contributions are threefold:
\begin{itemize}
	\item We mathematically formulate trustworthy uncertainty quantification for Bayesian regression.
	\item We propose \textsf{Trust-Bayes}, a novel optimization framework  for trustworthy uncertainty quantification.
	\item  We characterize the lower bounds of the probabilities of the ground truth being captured by the specified intervals and analyze the sample complexity with respect to the feasible probability for trustworthy uncertainty quantification.
\end{itemize}
We conduct Monte Carlo simulation and consider a case study using GPR for verification of trustworthy uncertainty quantification by \textsf{Trust-Bayes} and for comparison against \textsf{Meta-prior} \cite{lew2022safe,harrison2020meta,rothfuss2021pacoh} for its necessary.

{\em Notation.} Let $P_x\Big(\mathcal{E}\Big)$ return the probability of event $\mathcal{E}$ with respect to the probability measure of $x$ and $\mathbb{E}_x[\cdot]$ return the expected value with respect to the probability measure of $x$. Define indicator function $\bm{1}[\mathcal{E}]=1$ if event $\mathcal{E}$ is true and $\bm{1}[\mathcal{E}]=0$ otherwise. 
\section{Problem formulation}
{\em Observation model.}
Consider a distribution of  functions $\mathcal{P}_f$, where each unknown  function $f^i\sim\mathcal{P}_f$, $f^i:\mathcal{R}^{n_x}\to\mathbb{R}$,  can be observed as
\begin{align}
	y^i_t=f^i(x^i_t),
\end{align}
where $x^i_t\in\mathcal{X}\subset\mathcal{R}^{n_x}$ is the $t$th input to function $f^i$, $n_x$ is the dimensionality of $x_t$, and $y^i_t\in\mathbb{R}$ is the corresponding output. In this paper, we consider noiseless observation or the noise is inherent in $f^i$, which is a typical model for system identification \cite{fisac2019general}\cite{brunton2016discovering}\cite{dhiman2021control}. Here, $f^i$ can be different dynamic models of a system when operating in different environments, such as the dynamics of an autonomous car when operating in different weather conditions.

{\em Regression with uncertainty quantification.}
Denote the dataset of observations for function $f^i$ as $\mathcal{D}^i_{tr}\triangleq\{(y^i_t,x^i_t)\}_{t=1}^{t^i_{tr}}$, where $t^i_{tr}$ depends on $f^i$. Denote prior mean function $m:\mathcal{R}^{n_x}\to\mathbb{R}$,  prior covariance function $k:\mathcal{R}^{n_x}\times\mathcal{R}^{n_x}\to\mathbb{R}_{>0}$ which is positive semidefinite. Given dataset $\mathcal{D}^i_{tr}$, we denote the corresponding posterior mean function $\mu_{m,k}^i:\mathcal{R}^{n_x}\to\mathbb{R}$ and posterior standard deviation function $\sigma^i_{k}:\mathcal{R}^{n_x}\to\mathbb{R}$, which are output from some algorithm $\mathcal{ALG}(m,k,\mathcal{D}^i_{tr})$. The goal of this paper is to identify prior functions $m$ and $k$ which minimize some objective function $J(m,k)$ and meanwhile capture the ground truth $f^i(x)$ by intervals 
\begin{align*}
\mathcal{I}_{m,k}(x)&\triangleq[m(x)-q\sqrt{k(x,x)}, m(x)+q\sqrt{k(x,x)}]\\
\mathcal{I}^i_{m,k}(x)&\triangleq[\mu^i_{m,k}(x)-q\sigma^i_{k}(x), \mu^i_{m,k}(x)+q\sigma^i_{k}(x)]
\end{align*}
  with a pre-specified probability at least $1-\delta$, $\delta\in[0,1]$ for any $x\in\mathcal{X}$. Notice that constants $q$ and $\delta$ are fixed a priori, and they do not need to satisfy the relation between reliability factor and level-of-confidence in the settings of confidence interval. Formally, 
the above specifications for regression with uncertainty quantification is formulated as the optimization problem below
\begin{subequations}\label{problem}
\begin{align}
		\min_{m,k}&~J(m,k)\label{prob: obj}\\
	\text{s.t.}~& P_{f^i,x}\Big(f^i(x)\in\mathcal{I}_{m,k}(x)\Big)\geqslant1-\delta\label{prob: prior}\\
	&P_{f^i,x}\Big(f^i(x)\in\mathcal{I}^i_{m,k}(x)\Big)\geqslant1-\delta\label{prob: posterior}\\
	&\mu^i_{m,k},\sigma^i_{k}=\mathcal{ALG}(m,k,\mathcal{D}^i_{tr}).\label{prob: alg}
\end{align}
\end{subequations}
Constraints \eqref{prob: prior} and \eqref{prob: posterior} aim to ensure trustworthy uncertainty quantification, and one example of the objective function $J(m,k)$ can be negative marginal log likelihood (MLL) \cite{williams2006gaussian} evaluated over a dataset as in \cite{lew2022safe}\cite{harrison2020meta}. Notice  that in most cases given an arbitrary function $m(\cdot)$, intervals $\mathcal{I}_{m,k}(\cdot)$ and $\mathcal{I}^i_{m,k}(\cdot)$ can be arbitrarily large, and hence constraints \eqref{prob: prior} and \eqref{prob: posterior}  can always be satisfied, by choosing function $k$ such that $k(x,x)$ is sufficiently large if $\mathcal{ALG}$ follows a Bayesian inference framework, e.g., the prior distribution and the likelihood function are Gaussian \cite{bishop2006pattern}.

We do not assume any additional structure on  distribution $\mathcal{P}_f$ or on function $f^i$, e.g., $f^i\sim\mathcal{GP}(m,k)$, as in \cite{srinivas2012information}\cite{fisac2019general}\cite{yuan2024lightweight}. Instead, we assume we have access to a meta dataset $\mathcal{D}^{meta}\triangleq\{\mathcal{D}^i\}_{i=1}^n$, where $\mathcal{D}^i$ contains the observations of $f^i\sim\mathcal{P}_f$  i.i.d. drawn offline, which can be potentially used to estimate and optimize $J(m,k)$ as well as the left hand sides of \eqref{prob: prior} and \eqref{prob: posterior}. 
For testing, we consider functions $f^j\sim\mathcal{P}_f$, which  are not necessarily observed in $\mathcal{D}^{meta}$. 
\begin{remark}
{\em (Motivation of formulation \eqref{problem}).} Problem \eqref{problem} is motivated by the problems of safe learning/exploration \cite{fisac2019general} \cite{yuan2022dslap}\cite{lew2022safe}\cite{berkenkamp2023bayesian}\cite{wang2018safe}, where the system is required to safely explore and online learn about an unknown environment using the  data collected along the system's operation. During the early stage of exploration where there are only a few (or even no) collected data, the control of the system is mainly based on the prior knowledge, i.e., $(m,k)$, of the environment. Therefore, it is crucial that the selected prior is trustworthy such that  the ground truth of function $f^i$ can be captured a priori by $\mathcal{I}_{m,k}$, to ensure system safety  when the system just starts to explore the environment and no data is collected,  and a posteriori by $\mathcal{I}^i_{m,k}$ after collecting a (small) amount of data of $f^i$. 
$\hfill\blacksquare$
\end{remark}

{\em Parameterization.}
Note that Problem \eqref{problem} is a functional optimization problem and can be hard to solve in general. To make the problem tractable, we approximate the spaces of $m$ and $k$ using spaces of parameterized functions, e.g.,  neural networks and finite-dimensional basis functions. Specifically, we consider parameterized prior mean function $m_\theta$, with parameters $\theta\in\mathbb{R}^{n_\theta}$, e.g., the weights of a deep neural network, and parameterized prior mean function $k_\phi$, with parameters $\phi\in\mathbb{R}^{n_\phi}$.  Then Problem \eqref{problem} can be rewritten as 
\begin{subequations}\label{problem: param}
	\begin{align}
		\min_{\theta,\phi}&~J(\theta,\phi)\label{prob param: obj}\\
		\text{s.t.}~& P_{f^i,x}\Big(f^i(x)\in\mathcal{I}_{\theta,\phi}(x)\Big)\geqslant1-\delta\label{prob param: prior}\\
		&P_{f^i,x}\Big(f^i(x)\in\mathcal{I}^i_{\theta,\phi}(x)\Big)\geqslant1-\delta\label{prob param: posterior}\\
		&\mu^i_{\theta,\phi},\sigma^i_{\phi}=\mathcal{ALG}(m_\theta,k_\phi,\mathcal{D}^i_{tr}),
	\end{align}
\end{subequations}
with the terms rewritten accordingly.
In particular, we require $k_\phi$ to be scalable, i.e.,  $k_\phi\triangleq \phi_1\kappa_{\phi_2}$, where $\phi_1\in\mathbb{R}_{>0}$, $\phi_2\in\mathbb{R}^{n_\phi-1}$, $\phi=[\phi_1,\phi_2]$,  and $\kappa_{\phi_2}:\mathcal{X}\times\mathcal{X}\to[0,1]$ is a covariance function, such that the sizes of the intervals $\mathcal{I}_{\theta,\phi}(x)$ and $\mathcal{I}^i_{m,k}(x)$ can be arbitrarily large by increasing $\phi_1$, if  $\mathcal{ALG}$ follows some Bayesian inference frameworks such as GPR \cite{williams2006gaussian}.  Similar to \eqref{problem}, this implies that there always exists $\phi$ which satisfies \eqref{prob param: prior} and \eqref{prob param: posterior}, and hence Problem \eqref{problem: param} is always feasible.  One example of covariance function which satisfies the above requirement is  the widely-used square-exponential kernel $k(x,x')=\sigma_f^2\exp(-\frac{\|x-x'\|^2}{2\ell})+\sigma^2_e\delta_{xx'}$.

\section{The \textsf{Trust-Bayes} framework}
In this section, since distributions $\mathcal{P}_f$ and possibly $\mathcal{P}_x$ are unknown, we introduce the framework \textsf{Trust-Bayes}, to solve problem \eqref{problem: param} leveraging the empirical estimates from meta dataset $\mathcal{D}^{meta}$.

Define 0-1 loss functions
\begin{align*}
c^i_1(x)&\triangleq\bm{1}[f^i(x)\in\mathcal{I}_{\theta,\phi}(x)]\\
c^i_2(x)&\triangleq\bm{1}[f^i(x)\in\mathcal{I}^i_{\theta,\phi}(x)].
\end{align*}
For each $\mathcal{D}^i\in\mathcal{D}^{meta}$, it is split into  $\mathcal{D}^i_{tr}\subset\mathcal{D}^i$ for obtaining the posterior functions $(\mu^i_{\theta,\phi},\sigma^i_{k})$ for predictions and an evaluation dataset $\mathcal{D}^i_{eval}\triangleq \mathcal{D}^i\setminus\mathcal{D}^i_{tr}$ for evaluating the performances of the posterior functions and estimating \eqref{prob param: prior} and \eqref{prob param: posterior}. We write $\mathcal{D}^i_{eval}=\{(y^i_t,x^i_t)\}_{t=1}^{t^i_{eval}}$ and $\mathcal{D}^i_{tr}=\{(y^i_t,x^i_t)\}_{t=1}^{t^i_{tr}}$.For any $\gamma\in(0,0.5]$, define
\begin{align*}
	&p_1(\gamma)\triangleq \Big(1-2\gamma\Big)\cdot\Big(\frac{1}{n}\sum_{i=1}^n\frac{1}{t^i_{eval}}\sum_{t=1}^{t^i_{eval}} c^i_1(x^i_t)\nonumber\\
	&\qquad\qquad-\sqrt{\frac{\log(2/\gamma)\sum_{i=1}^n\frac{1}{t^i_{eval}}}{2n^2}}-\sqrt{\frac{\log (2/\gamma)}{2n}}\Big),\\
	&p_2(\gamma)\triangleq \Big(1-2\gamma\Big)\cdot\Big(\frac{1}{n}\sum_{i=1}^n\frac{1}{t^i_{eval}}\sum_{t=1}^{t^i_{eval}} c^i_2(x^i_t)\nonumber\\
	&\qquad\qquad-\sqrt{\frac{\log(2/\gamma)\sum_{i=1}^n\frac{1}{t^i_{eval}}}{2n^2}}-\sqrt{\frac{\log (2/\gamma)}{2n}}\Big).
\end{align*}
Then the following theorem characterizes the lower bounds of 
$P_{f^i,x}\Big(f^i(x)\in\mathcal{I}_{\theta,\phi}(x)\Big)$ and $P_{f^i,x}\Big(f^i(x)\in\mathcal{I}^i_{\theta,\phi}(x)\Big)$.
\begin{theorem}\label{thm: PP}
Suppose $\{f^i\}_{i=1}^n$ are sampled i.i.d. from a latent distribution $\mathcal{P}_f$.  Suppose the training dataset $\mathcal{D}^i_{tr}$ for each function $f^i$ is generated through a latent conditional distribution $\mathcal{P}_{\mathcal{D}^i_{tr}\mid f^i}$, i.e., $\mathcal{D}^i_{tr}\sim\mathcal{P}_{\mathcal{D}^i_{tr}\mid f^i}$. Suppose for each $f^i$, $\{x^i_t\}_{t=1}^{t^i_{eval}}$ are sampled i.i.d. from a latent function $\mathcal{P}_x$.  
 Then the following inequalities hold:
\begin{align*}
&P_{f^i,x}\Big(f^i(x)\in\mathcal{I}_{\theta,\phi}(x)\Big)\geqslant \max_{\gamma\in(0,0.5]}p_1(\gamma),
\\
&P_{f^i,x}\Big(f^i(x)\in\mathcal{I}^i_{\theta,\phi}(x)\Big)\geqslant \max_{\gamma\in(0,0.5]}p_2(\gamma).
\end{align*}
$\hfill\blacksquare$
\end{theorem}
The proof of the theorem is deferred to Section \ref{sec: proof}. The lower bounds in Theorem \ref{thm: PP} indicates that $P_{f^i,x}\Big(f^i(x)\in\mathcal{I}_{\theta,\phi}(x)\Big)$ and $P_{f^i,x}\Big(f^i(x)\in\mathcal{I}^i_{\theta,\phi}(x)\Big)$ can be lower bounded by their corresponding empirical estimates of  (i.e., $\frac{1}{n}\sum_{i=1}^n\frac{1}{t^i_{eval}}\sum_{t=1}^{t^i_{eval}} c^i_1(x^i_t)$ and $\frac{1}{n}\sum_{i=1}^n\frac{1}{t^i_{eval}}\sum_{t=1}^{t^i_{eval}} c^i_2(x^i_t)$) using the evaluation datasets $\mathcal{D}^i_{eval}$, $i=1,\cdots,n$, in the meta training dataset $\mathcal{D}^{meta}$, and the error terms diminish by $t^i_{eval}$ and $n$, the sizes of $\mathcal{D}^i_{eval}$ and the number of functions in $\mathcal{D}^{meta}$. This provides empirical underestimates of $P_{f^i,x}\Big(f^i(x)\in\mathcal{I}_{\theta,\phi}(x)\Big)$ and $P_{f^i,x}\Big(f^i(x)\in\mathcal{I}^i_{\theta,\phi}(x)\Big)$ but sufficient verification for constraints \eqref{prob param: prior} and \eqref{prob param: posterior}.

By the lower bounds in the theorem above, we can approximate Problem \eqref{problem: param} with
\begin{subequations}\label{problem: param approx}
	\begin{align}
		\min_{\theta,\phi}&~J(\theta,\phi)\label{prob param approx: obj}\\
		\text{s.t.}~& \max_{\gamma\in(0,0.5]}p_1(\gamma)\geqslant1-\delta,\label{prob param approx: prior}\\
		&\max_{\gamma\in(0,0.5]}p_2(\gamma)\geqslant1-\delta,\label{prob param approx: posterior}\\
		&\mu^i_{\theta,\phi},\sigma^i_{\phi}=\mathcal{ALG}(m_\theta,k_\phi,\mathcal{D}^i_{tr}),\label{prob param approx: bayes}
	\end{align}
\end{subequations}
which is our proposed formulation of \textsf{Trust-Bayes}. 

\begin{remark}
	{\em (Feasibility and sample complexity).} As discussed  below Problem \eqref{problem: param}, the sizes of intervals $\mathcal{I}_{\theta,\phi}(x)$ and $\mathcal{I}_{\theta,\phi}^i(x)$ can be arbitrarily large by choosing $\phi$ accordingly. This implies that $\frac{1}{n}\sum_{i=1}^n\frac{1}{t^i_{eval}}\sum_{t=1}^{t^i_{eval}} c^i_j(x_t)$, $j\in\{1,2\}$, can be as large as one. Therefore, by the forms of \eqref{prob param approx: prior} and \eqref{prob param approx: posterior}, the feasibility of \eqref{problem: param approx} can be checked by verifying whether
	\begin{align}\label{ineq: feas check}
	\max_{\gamma\in(0,0.5]}& (1-2\gamma)\Big(1-\sqrt{\frac{\log(2/\gamma)\sum_{i=1}^n\frac{1}{t^i_{eval}}}{2n^2}}-\sqrt{\frac{\log (2/\gamma)}{2n}}\Big)\nonumber\\
	&\geqslant1-\delta
	\end{align}
	holds for a given $\delta$, which can be done through numerically solving the left hand side using, e.g., gradient ascent. Note that for any $\gamma\in(0,0.5]$, $\sqrt{\frac{\log(2/\gamma)\sum_{i=1}^n\frac{1}{t^i_{eval}}}{2n^2}}$ and $\sqrt{\frac{\log (2/\gamma)}{2n}}$ can be arbitrarily small by increasing sample sizes $t^i_{eval}$ and $n$. Therefore,  inequality \eqref{ineq: feas check} provides the sample complexity for the lower bound of feasible $\delta$ and implies that  the freedom of selecting $\delta$ can be increased by increasing  $t^i_{eval}$ and $n$.
	 Note that \eqref{prob param approx: prior} and \eqref{prob param approx: posterior} approximate the probability constraints \eqref{prob param: prior} and \eqref{prob param: posterior} by underestimating the probabilities with the corresponding empirical estimates.
	 This is reminiscent of the no-free-lunch property between estimation error and the number of data in statistical learning theory \cite{vapnik2013nature}. $\hfill\blacksquare$
\end{remark}

\section{Proof of Theorem \ref{thm: PP}}\label{sec: proof}

The proof of the theorem leverages several properties of sub-Gaussian random variables defined as follows.
\begin{definition}
A random variable $X$ with mean $\mu=\mathbb{E}[X]$ is sub-Gaussian if there is a positive number $\sigma$ such that $\mathbb{E}[e^{\lambda(X-\mu)}]\leqslant e^{\sigma^2\lambda^2/2}$ for all $\lambda\in\mathbb{R}$. $\hfill\blacksquare$
\end{definition}
Based on the above definition, we can denote a sub-Gaussian random variable $X$ as $X\sim \textsf{sub-Gauss}(\mu,\sigma)$
The properties leveraged in the proof are included below for completeness, and they can be found in \cite{vershynin2018high}.
\begin{lemma}\label{lemma: bounded sub gauss}
	{\em (Bounded random variables are sub-Gaussian).}
A random variable $X$ with mean $\mu=\mathbb{E}[X]$ bounded as $a\leqslant X\leqslant b$ is sub-Gaussian with $\sigma^2=(\frac{b-a}{2})^2$. $\hfill\blacksquare$
\end{lemma}
\begin{lemma}\label{lemma: sub gauss super}
{\em (Preservation under superposition).} If $X_1\sim \textsf{sub-Gauss}(\mu_1,\sigma_1^2)$ and $X_2\sim \textsf{sub-Gauss}(\mu_2,\sigma_2^2)$, then $X_1+X_2\sim \textsf{sub-Gauss}(\mu_1+\mu_2, \sigma_1^2+\sigma_2^2)$ and $aX_1\sim \textsf{sub-Gauss}(a\mu_1,a^2\sigma_1^2)$.
$\hfill\blacksquare$
\end{lemma}
\begin{lemma}\label{lemma: Hoeffding's inequality}
	{\em (Hoeffding's inequality).} Suppose that the variables $X_i\sim \textsf{sub-Gauss}(\mu,\sigma^2)$, $i=1,\cdots,n$, are independent. Then for all $\epsilon\geqslant0$, we have
		$$
	P\Big(|X-\mu|\geqslant \epsilon\Big)\leqslant 2\exp\Big(-\frac{\epsilon^2}{2\sigma^2}\Big). 
	$$ 
$\hfill\blacksquare$
\end{lemma}


Before proving the theorem, we first introduce the following expectations and empirical estimates:
\begin{align*}
	C_1&\triangleq\mathbb{E}_{f^i,x}[c^i_1(x)]=P_{f^i,x}\Big(f^i(x)\in\mathcal{I}_{\theta,\phi}(x)\Big)\\
	C_2&\triangleq\mathbb{E}_{f^i,x}[c^i_2(x)]=P_{f^i,x}\Big(f^i(x)\in\mathcal{I}^i_{\theta,\phi}(x)\Big)\\
	C^i_{1}&\triangleq \mathbb{E}_{x}[c^i_1(x_t)\mid f^i], \quad C^i_{2}\triangleq \mathbb{E}_{x}[c^i_2(x_t)\mid f^i],\\
	\hat{C}^i_{1}&\triangleq \frac{1}{t^i_{eval}}\sum_{t=1}^{t^i_{eval}} c^i_1(x^i_t), \quad\hat{C}^i_{2}\triangleq \frac{1}{t^i_{eval}}\sum_{t=1}^{t^i_{eval}} c^i_2(x^i_t).
\end{align*}
Then the following lemma characterizes the lower bounds for expectations $C_1$ and $C_2$ in terms of empirical estimates $\hat{C}^i_{1}$ and $\hat{C}^i_{2}$.
\begin{lemma}\label{lemma: C1 C2 leq}
Each of the followings hold with probability at least $1-2\gamma$:
\begin{align*}
		C_{1}&\geqslant\frac{1}{n}\sum_{i=1}^n\hat{C}^i_{1}-\sqrt{\frac{\log(2/\gamma)\sum_{i=1}^n\frac{1}{t^i_{eval}}}{2n^2}}-\sqrt{\frac{\log (2/\gamma)}{2n}},\\
	C_{2}&\geqslant\frac{1}{n}\sum_{i=1}^n\hat{C}^i_{2}-\sqrt{\frac{\log(2/\gamma)\sum_{i=1}^n\frac{1}{t^i_{eval}}}{2n^2}}-\sqrt{\frac{\log (2/\gamma)}{2n}}.
\end{align*}

{\bf Proof:}
By the definitions of $c^i_1$ and $C^i_1$, we have $C^i_1\in[0,1]$ and $\mathbb{E}_{f^i}[C^i_1]=C_1$. Then Lemma \ref{lemma: bounded sub gauss} renders that $C^i_1\sim \textsf{sub-Gauss}(C_1,\frac{1}{4})$, and hence Lemma \ref{lemma: sub gauss super} renders that $\frac{1}{n}\sum_{i=1}^nC^i_1\sim  \textsf{sub-Gauss}(C_1,\frac{1}{4n})$. Therefore, by Lemma \ref{lemma: Hoeffding's inequality}, we have 	$
P\Big(|\frac{1}{n}\sum_{i=1}^nC^i_1-C_1|\geqslant \epsilon\Big)\leqslant 2\exp\Big(-2n\epsilon^2\Big). 
$  After some simple algebraic transformations, we have
\begin{align}\label{ineq: C 1}
	C_{1}\geqslant\frac{1}{n}\sum_{i=1}^nC^i_{1}-\sqrt{\frac{\log (2/\gamma)}{2n}}.
\end{align}
with probability at least $1-\gamma$.

Recall the definitions of $ c^i_1(x_t)$ and $C^i_{1}$.
Notice that $c^i_1(x_t)\in[0,1]$ for all $i=1,\cdots,n$ and $t=1,2,\cdots$. Then Lemma \ref{lemma: bounded sub gauss} renders that $c^i_1(x_t)\sim \textsf{sub-Gauss}(C^i_1,\frac{1}{4})$, and Lemma \ref{lemma: sub gauss super} renders that $\hat{C}^i_1=\frac{1}{t^i_{eval}}\sum_{t=1}^{t^i_{eval}}c^i_1(x_t)\sim \textsf{sub-Gauss}(C^i_1,\frac{1}{4t^i_{eval}}) $.
Furthermore, by Lemma \ref{lemma: sub gauss super}, we also have $\frac{1}{n}\sum_{i=1}^n\hat{C}^i_1\sim\textsf{sub-Gauss}(\frac{1}{n}\sum_{i=1}^nC^i_1, \frac{1}{n^2}\sum_{i=1}^n\frac{1}{4t^i_{eval}})$.
Then following similar logic to \eqref{ineq: C 1}, we have
\begin{align}\label{ineq: C 1 fi}
	\frac{1}{n}\sum_{i=1}^nC^i_1\geqslant\frac{1}{n}\sum_{i=1}^n\hat{C}^i_1-\sqrt{\frac{\log(2/\gamma)\sum_{i=1}^n\frac{1}{t^i_{eval}}}{2n^2}}.
\end{align}
with probability at least $1-\gamma$.
Combining \eqref{ineq: C 1 fi} with \eqref{ineq: C 1} renders
\begin{align*}
C_{1}\geqslant\frac{1}{n}\sum_{i=1}^n\hat{C}^i_{1}-\sqrt{\frac{\log(2/\gamma)\sum_{i=1}^n\frac{1}{t^i_{eval}}}{2n^2}}-\sqrt{\frac{\log (2/\gamma)}{2n}}
\end{align*}
with probability at least $(1-\gamma)^{2}\geqslant1-2\gamma$.
Following the same logic, we also have, with probability at least $1-2\gamma$,
\begin{align*}
	C_{2}\geqslant\frac{1}{n}\sum_{i=1}^n\hat{C}^i_{2}-\sqrt{\frac{\log(2/\gamma)\sum_{i=1}^n\frac{1}{t^i_{eval}}}{2n^2}}-\sqrt{\frac{\log (2/\gamma)}{2n}}.
\end{align*}
$\hfill\blacksquare$
\end{lemma}

The the proof of Theorem \ref{thm: PP} is given as follows.

Let $G_{1,\gamma}\triangleq\frac{1}{n}\sum_{i=1}^n\hat{C}^i_{1}-\sqrt{\frac{\log(2/\gamma)\sum_{i=1}^n\frac{1}{t^i_{eval}}}{2n^2}}-\sqrt{\frac{\log (2/\gamma)}{2n}}$. Then by Lemma \ref{lemma: C1 C2 leq}, we have
\begin{align}\label{ineq: P f in I1}
	&P_{f^i,x}\Big(f^i(x)\in\mathcal{I}_{\theta,\phi}(x)\Big)\nonumber\\
	&=P_{f^i,x}\Big(f^i(x)\in\mathcal{I}_{\theta,\phi}(x)| C_1\geqslant G_{\hat{C}^i_{1},\gamma}\Big)P\Big( C_1\geqslant G_{\hat{C}^i_{1},\gamma}\Big)\nonumber\\
	&~+P_{f^i,x}\Big(f^i(x)\in\mathcal{I}_{\theta,\phi}(x)| C_1< G_{\hat{C}^i_{1},\gamma}\Big)P\Big( C_1< G_{\hat{C}^i_{1},\gamma}\Big)\nonumber\\
	&\geqslant P_{f^i,x}\Big(f^i(x)\in\mathcal{I}_{\theta,\phi}(x)| C_1\geqslant G_{\hat{C}^i_{1},\gamma}\Big)P\Big( C_1\geqslant G_{\hat{C}^i_{1},\gamma}\Big)\nonumber\\
	&\geqslant\Big(\frac{1}{n}\sum_{i=1}^n\hat{C}^i_{1}-\sqrt{\frac{\log(2/\gamma)\sum_{i=1}^n\frac{1}{t^i_{eval}}}{2n^2}}-\sqrt{\frac{\log (2/\gamma)}{2n}}\Big)\nonumber\\
	&\qquad\cdot\Big(1-2\gamma\Big).
\end{align}
for any $\gamma\in(0,0.5]$. Following the same logic, we have
\begin{align}\label{ineq: P f in Ii1}
&P_{f^i,x}\Big(f^i(x)\in\mathcal{I}^i_{\theta,\phi}(x)\Big)\nonumber\\
&\geqslant\Big(\frac{1}{n}\sum_{i=1}^n\hat{C}^i_{2}-\sqrt{\frac{\log(2/\gamma)\sum_{i=1}^n\frac{1}{t^i_{eval}}}{2n^2}}-\sqrt{\frac{\log (2/\gamma)}{2n}}\Big))\nonumber\\
&\qquad\cdot\Big(1-2\gamma\Big).
\end{align}
for any $\gamma\in(0,0.5]$. To ensure the tightness of the bounds, we maximize the right hand sides of \eqref{ineq: P f in I1} and \eqref{ineq: P f in Ii1} with respect to $\gamma$.
 This completes the proof of the theorem. $\hfill\blacksquare$

\section{Case study}
In this section, we aim to verify whether the \textsf{Trust-Bayes} formulation in \eqref{problem: param approx} can provide trustworthy uncertainty quantification,  i.e., satisfying the constraints in \eqref{problem: param}, and whether it is necessary. 
In this case study, we consider GPR \cite{williams2006gaussian} as the algorithm $\mathcal{ALG}$. Then the prior predictive distribution for $f^i(x)$ is given by $\mathcal{N}(m_\theta(x), k_\phi(x,x))$ and the posterior predictive distribution is given by $\mathcal{N}(\mu^i_{\theta,\phi}(x),(\sigma^i_{\phi})^2(x))$, where
\begin{align*}
	&\mu^i_{\theta,\phi}(x)=m_\theta(x)+k_\phi(x,X^i)k_\phi^{-1}(X^i,X^i)(Y^i-m_\theta(X^i))\\
	&(\sigma^i_{\phi})^2(x)=k_\phi(x,x)+k_\phi(x,X^i)k_\phi^{-1}(X^i,X^i)k_\phi(X^i,x).
\end{align*}
where $X^i$ aggregates all the inputs $x^i_t$ in $\mathcal{D}^i$, $Y^i$ aggregates all the outputs. 	

{\em Experiment setup.} In this experiment,
 we let $x\in[0,1]$ and for each $f^i\in\mathcal{P}_{f}$,   $f^i(x)=d^ix^2+\sum_{m=1}^{10}\alpha^ia^i_m\sin(w^i_mx+\beta^i_m)+(1-\alpha^i)b^i_m\sin(u^i_mx+\beta^i_m)$, where
\begin{subequations}\label{eq: simulation functions}
\begin{align}
	a^i_m&\sim0.5\mathcal{N}(-20,5)+0.5\mathcal{N}(10,2)\\
	b^i_m&\sim0.5\mathcal{N}(-1,0.1)+0.5\mathcal{N}(1,0.1)\\
	w^i_m&\sim0.5\mathcal{N}(-10,10)+0.5\mathcal{N}(10,10)\\
	u^i_m&\sim0.5\mathcal{N}(-100,10)+0.5\mathcal{N}(100,10)\\
	\beta^i_m&\sim\mathcal{N}(0,1)\\
	d^i&\sim0.5\mathcal{N}(-10,1)+0.5\mathcal{N}(10,1)\\
	\alpha^i&\sim\mathcal{B}(0.5).
\end{align}
\end{subequations}
 $\mathcal{N}$ denotes normal distribution and $\mathcal{B}$ denotes Bernoulli distribution. 
For learning using GPR, we consider constant $m_\theta(x)=\theta$ for the prior mean function and kernel $k_\phi(x,x')=\phi_1^2 \exp(-\|x-x'\|^2\phi_2)$. For trustworthy uncertainty quantification, we select $\delta=0.1$ and $q_\delta=1.64$ for intervals $\mathcal{I}_{\theta,\phi}(x)$ and $\mathcal{I}^i_{\theta,\phi}(x)$. That is, we require the  predicted intervals with  90\% nominal confidence level, according to the Z-score table, to indeed include at least 90\% of the true values. Note that according to \eqref{eq: simulation functions}, there are parameters following Gaussian mixtures distributions and Bernoulli distribution, and therefore $f^i$ does not follow a Gaussian process. Furthermore,  each $f^i$  has a trend, and therefore $k_\phi(x,x')$ is not a suitable prior covariance function. The purpose of this setup is to demonstrate that \textsf{Trust-Bayes} is able to provide trustworthy uncertainty quantification even under such mis-specification of the prior distribution, which can happen when there is no much prior knowledge on the distribution and structure on the target function $f^i$.

{\em Training.}
For the training dataset $\mathcal{D}^{tr}$, we sample 2000 functions $f^i$, i.e., $n=2000$, following \eqref{eq: simulation functions}, and for each  corresponding training dataset $\mathcal{D}^i$ we uniformly sample over interval $[0,1]$ for inputs $t^i_{tr}=20$ for obtaining the posterior and $t^i_{eval}=100$ for evaluating the posterior for \eqref{prob param approx: obj} to \eqref{prob param approx: posterior}. 
Note that with the above specification of $\delta$, $n$, and $t^i_{eval}$, \eqref{ineq: feas check} holds and can be verified by plugging in $\gamma=0.001$.
For the objective function $J(\theta,\phi)$, we consider negative MLL over the whole meta dataset  $\mathcal{D}^{tr}$ as in \cite{lew2022safe,harrison2020meta,rothfuss2021pacoh}.

{\em Testing.} Notice that the probabilities in \eqref{prob param: prior} and \eqref{prob param: posterior} can be challenging to exactly obtain. As an estimation, the trained hyperparameters $(\theta,\phi)$ are tested against a dataset composed of  $10,000$ functions $f^i$ randomly sampled over \eqref{eq: simulation functions} with each $f^i$ has testing $10,000$ inputs $x$ uniformly sampled over $[0,1]$. The corresponding posterior distribution is obtained after observing each function over 20 inputs without further tuning the hyperparameters.

%
%

{\em Comparison.}
Our results are compared against the \textsf{Meta-prior} method used in \cite{lew2022safe,harrison2020meta,rothfuss2021pacoh}, where the hyperparameters in the prior distributions are meta-trained using negative MLL. Note that the Bayesian regression approach used in \cite{lew2022safe}\cite{harrison2020meta} is a special case of GPR \cite{williams2006gaussian}. The difference between \textsf{Trust-Bayes} and \textsf{Meta-prior} in this case study is the addition of \eqref{prob param approx: prior} and \eqref{prob param approx: posterior}.

\begin{figure}[t]
	\centering
	\begin{subfigure}[b]{0.235\textwidth}	
		\includegraphics[width=1.\textwidth]{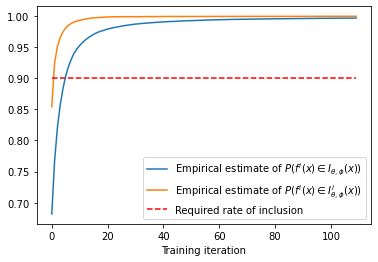}
		\caption{Trust-Bayes}
	\end{subfigure}	
	\begin{subfigure}[b]{0.235\textwidth}	
		\centering
		\includegraphics[width=1.\textwidth]{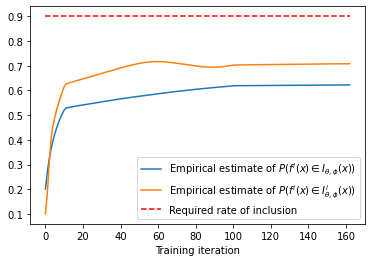}
		\caption{Meta-prior}
	\end{subfigure}
	\caption{Comparison over empirical estimates of $P_{f^i,x}\Big(f^i(x)\in\mathcal{I}_{\theta,\phi}(x)\Big)$ and $P_{f^i,x}\Big(f^i(x)\in\mathcal{I}^i_{\theta,\phi}(x)\Big)$ by the evaluation dataset  $\{\mathcal{D}^{i}_{eval}\}_{i=1}^n$ between Trust-Bayes and Meta-prior during meta-training}
	\label{fig: estimates}
\end{figure}

{\em Results.}
Figure \ref{fig: estimates} shows the convergence of $\frac{1}{n}\sum_{i=1}^n\frac{1}{t^i_{eval}}\sum_{t=1}^{t^i_{eval}} c^i_1(x^i_t)$ and $\frac{1}{n}\sum_{i=1}^n\frac{1}{t^i_{eval}}\sum_{t=1}^{t^i_{eval}} c^i_2(x^i_t)$, the empirical estimates of $P_{f^i,x}\Big(f^i(x)\in\mathcal{I}_{\theta,\phi}(x)\Big)$ and $P_{f^i,x}\Big(f^i(x)\in\mathcal{I}^i_{\theta,\phi}(x)\Big)$, respectively, by the evaluation dataset  $\{\mathcal{D}^{i}_{eval}\}_{i=1}^n$ in the meta dataset $\mathcal{D}^{meta}$. From Figure \ref{fig: estimates}, we can see that the training of \textsf{Meta-prior} converges when the empirical estimates are still at a level much lower than the required rate of inclusion $1-\delta=0.9$. In contrast, the training of \textsf{Trust-Bayes} converges much faster and beyond the required rate of inclusion.

Table \ref{table: inclusion} provides a comparison between the  $(\theta,\phi)$ trained till convergence by \textsf{Trust-Bayes} and that by \textsf{Meta-prior} over the estimated $P_{f^i,x}\Big(f^i(x)\in\mathcal{I}_{\theta,\phi}(x)\Big)$, $P_{f^i,x}\Big(f^i(x)\in\mathcal{I}^i_{\theta,\phi}(x)\Big)$,  their empirical estimates by $\mathcal{D}^i_{eval}$ as well as the mean-squared error (MSE) for the posterior predictions over the testing dataset. 
From Table   \ref{table: inclusion}, we can see that \textsf{Trust-Bayes} performs better than \textsf{Meta-prior} over all the metrics. Specifically,  $P_{f^i,x}\Big(f^i(x)\in\mathcal{I}_{\theta,\phi}(x)\Big)$ and $P_{f^i,x}\Big(f^i(x)\in\mathcal{I}^i_{\theta,\phi}(x)\Big)$ are higher than the required inclusion rate (i.e, 0.9) when $(\theta,\phi)$ are trained by \textsf{Trust-Bayes}, which verifies that \textsf{Trust-Bayes} can provide trustworthy uncertainty quantification. In contrast,  when  $(\theta,\phi)$ are trained by \textsf{Meta-prior},  the rates of inclusion are much lower than the required inclusion rate  although the intervals are constructed with 90\% nominal  level of confidence.  This comparison with \textsf{Meta-prior} shows the necessary of using \textsf{Trust-Bayes} to provide trustworthy uncertainty quantification when it is uncertain whether the prior is properly specified or not.
\begin{table}[t]
	\begin{center}
		\begin{tabular}{ |c|c|c| } 
			\hline
			&Trust-Bayes&Meta-prior\\
			\hline	
			$\frac{1}{n}\sum_{i=1}^n\frac{1}{t^i_{eval}}\sum_{t=1}^{t^i_{eval}} c^i_1(x^i_t)$ & 0.997 & 0.623\\ 
			\hline
			$\frac{1}{n}\sum_{i=1}^n\frac{1}{t^i_{eval}}\sum_{t=1}^{t^i_{eval}} c^i_2(x^i_t)$ &0.999  & 0.706 \\ 
			\hline
			$P_{f^i,x}\Big(f^i(x)\in\mathcal{I}_{\theta,\phi}(x)\Big)$ & 0.996  &0.630\\ 
			\hline
			$P_{f^i,x}\Big(f^i(x)\in\mathcal{I}^i_{\theta,\phi}(x)\Big)$ &0.999  & 0.708\\ 
			\hline
			MSE& 91.42&125.15\\
			\hline
		\end{tabular}
		\caption{Comparisons over empirical and expected prior/posterior inclusions  between \textsf{Trust-Bayes} and \textsf{Meta-prior}}
		\label{table: inclusion}
	\end{center}
	
\end{table}

\begin{figure}
	\centering
		\begin{subfigure}[b]{0.3\textwidth}	
		\includegraphics[width=1.\textwidth]{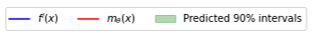}	
	\end{subfigure}	
		\begin{subfigure}[b]{0.235\textwidth}	
	\includegraphics[width=1.1\textwidth]{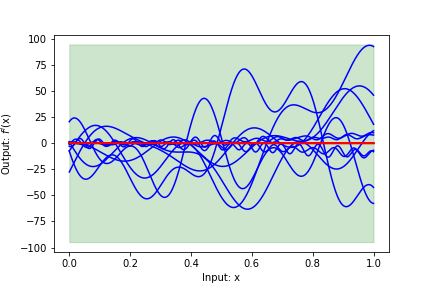}
	\caption{Trust-Bayes}
		\end{subfigure}	
			\begin{subfigure}[b]{0.235\textwidth}	
					\centering
				\includegraphics[width=1.1\textwidth]{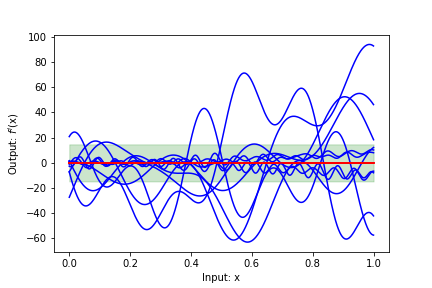}
				\caption{Meta-prior}
						\end{subfigure}
			\caption{Ten random functions drawn from \eqref{eq: simulation functions} and comparison between prior predictions by Trust-Bayes and Meta-prior}
			\label{fig: prior}
\end{figure}

\begin{figure*}
	\centering
	\begin{subfigure}[b]{0.3\textwidth}	
	\includegraphics[width=1.\textwidth]{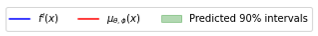}	
\end{subfigure}
	\begin{subfigure}[b]{1.0\textwidth}	
		\includegraphics[width=0.195\textwidth]{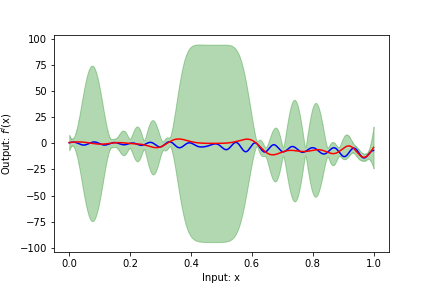}
		\includegraphics[width=0.195\textwidth]{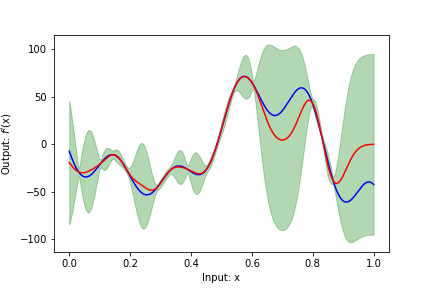}
		\includegraphics[width=0.195\textwidth]{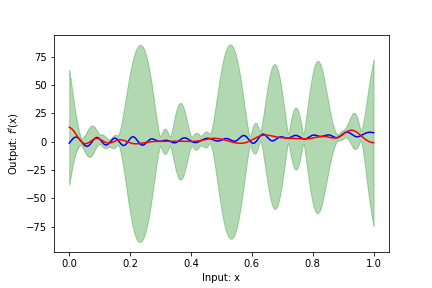}
		\includegraphics[width=0.195\textwidth]{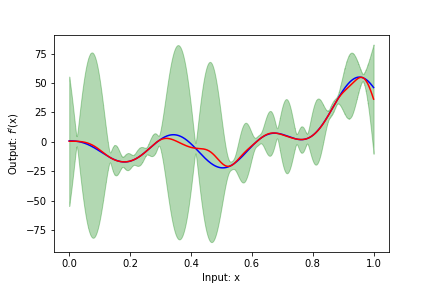}
		\includegraphics[width=0.195\textwidth]{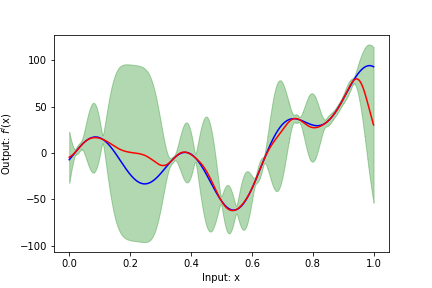}
		
	\end{subfigure}
	\begin{subfigure}[b]{1.0\textwidth}
		\includegraphics[width=0.195\textwidth]{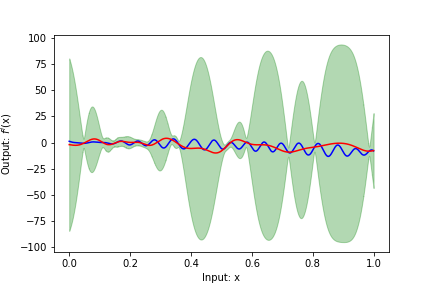}
	\includegraphics[width=0.195\textwidth]{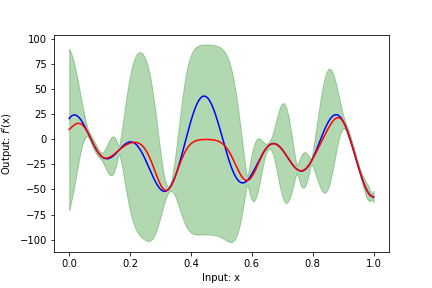}
	\includegraphics[width=0.195\textwidth]{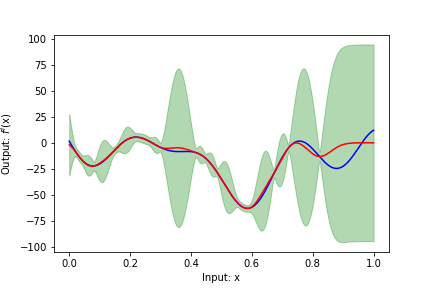}
	\includegraphics[width=0.195\textwidth]{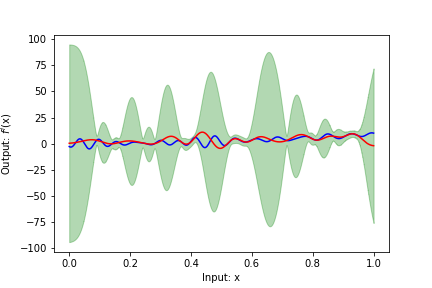}
	\includegraphics[width=0.195\textwidth]{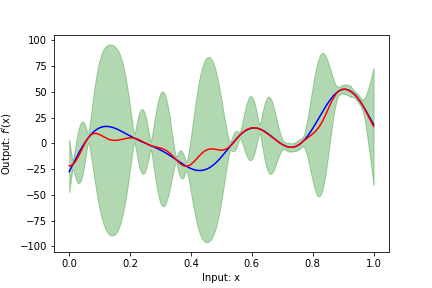}
		\caption{Posterior predictions over ten sample functions by \textsf{Trust-Bayes}}
	\label{fig: post by TB}
	\end{subfigure}

	\begin{subfigure}[b]{1.0\textwidth}	
		\includegraphics[width=0.195\textwidth]{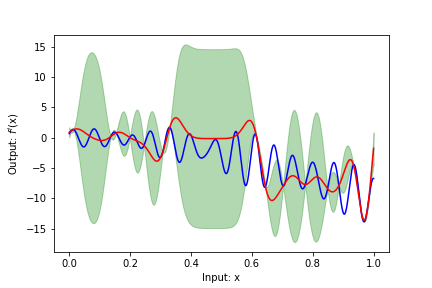}
		\includegraphics[width=0.195\textwidth]{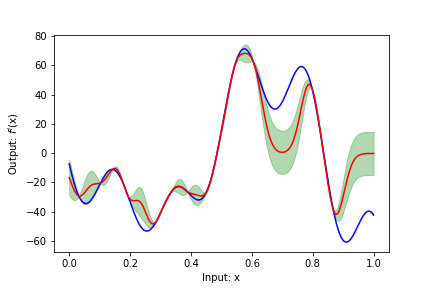}
		\includegraphics[width=0.195\textwidth]{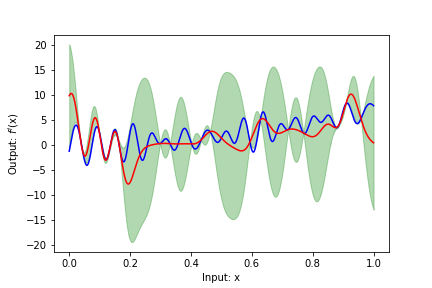}
		\includegraphics[width=0.195\textwidth]{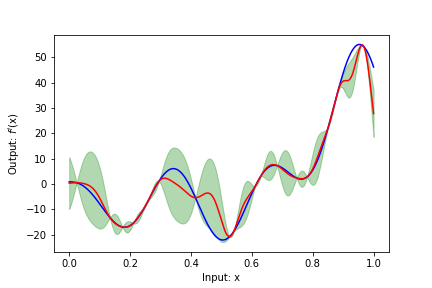}
		\includegraphics[width=0.195\textwidth]{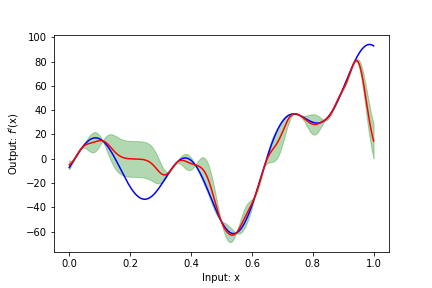}
		
	\end{subfigure}
	\begin{subfigure}[b]{1.0\textwidth}
		\includegraphics[width=0.195\textwidth]{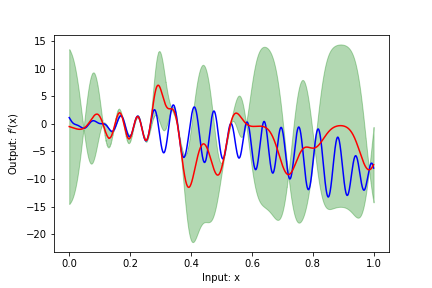}
		\includegraphics[width=0.195\textwidth]{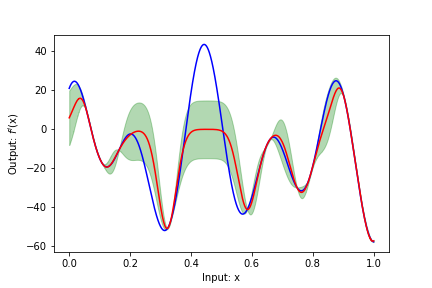}
		\includegraphics[width=0.195\textwidth]{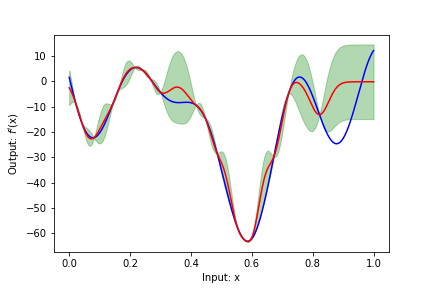}
		\includegraphics[width=0.195\textwidth]{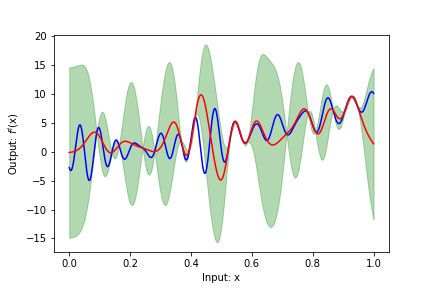}
		\includegraphics[width=0.195\textwidth]{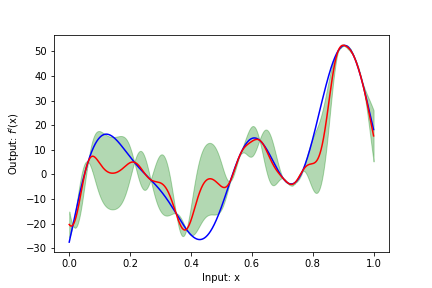}
			\caption{Posterior predictions over ten sample functions by \textsf{Meta-prior}}
		\label{fig: post by meta prior}
	\end{subfigure}
	\caption{Comparison between posterior predictions  by \textsf{Trust-Bayes} and \textsf{Meta-prior} over ten random functions from \eqref{eq: simulation functions}}
\label{fig: post}
\end{figure*}

Figure \ref{fig: prior}  provides a visual comparison over the inclusions of the true values of $f^i(x)$ by the prior 90\% confidence intervals  $\mathcal{I}_{\theta,\phi}(x)$ trained using \textsf{Trust-Bayes} and \textsf{Meta-prior}, respectively, and Figure \ref{fig: post}  provides a comparison over the inclusions by the posterior 90\% confidence intervals  $\mathcal{I}^i_{\theta,\phi}(x)$. Ten functions are sampled from \eqref{eq: simulation functions} to provide a visualization of the distribution of the functions. From these two figures, we can see that the 90\% intervals  generated by \textsf{Trust-Bayes} are larger and are able to capture most of the values of the functions including those with larger variation, whereas \textsf{Meta-prior} generates smaller intervals and fails to capture the values of the functions with larger variation.

\section{Conclusion}
We consider trustworthy uncertainty quantification in Bayesian regression problems.
We formulate trustworthy uncertainty quantification as constraints on capturing the ground truths of the function by intervals depending on the prior and posterior distributions with a pre-specified probability.
We propose, \textsf{Trust-Bayes}, a Bayesian meta learning framework which is cognizant of trustworthy uncertainty quantification without explicit assumptions on the model/distribution of the functions.  We characterize the lower bounds of the probabilities of the ground truth being captured by the specified intervals in terms of the empirical estimates  and analyze the sample complexity with respect to the feasible pre-specified probability  for trustworthy uncertainty quantification.   Monte Carlo simulation is conducted for evaluation and comparison through a case study using GPR, which verifies the proposed framework and demonstrates the necessary of  \textsf{Trust-Bayes} for trustworthy uncertainty quantification when the prior is not necessarily correctly specified.

\bibliographystyle{IEEEtran}
\bibliography{Biblio-Database1}

\begin{thebibliography}{10}
\providecommand{\url}[1]{#1}
\csname url@samestyle\endcsname
\providecommand{\newblock}{\relax}
\providecommand{\bibinfo}[2]{#2}
\providecommand{\BIBentrySTDinterwordspacing}{\spaceskip=0pt\relax}
\providecommand{\BIBentryALTinterwordstretchfactor}{4}
\providecommand{\BIBentryALTinterwordspacing}{\spaceskip=\fontdimen2\font plus
\BIBentryALTinterwordstretchfactor\fontdimen3\font minus
  \fontdimen4\font\relax}
\providecommand{\BIBforeignlanguage}[2]{{%
\expandafter\ifx\csname l@#1\endcsname\relax
\typeout{** WARNING: IEEEtran.bst: No hyphenation pattern has been}%
\typeout{** loaded for the language `#1'. Using the pattern for}%
\typeout{** the default language instead.}%
\else
\language=\csname l@#1\endcsname
\fi
#2}}
\providecommand{\BIBdecl}{\relax}
\BIBdecl

\bibitem{mitchell1997machine}
T.~M. Mitchell, ``Machine learning,'' 1997.

\bibitem{williams2006gaussian}
C.~K. Williams and C.~E. Rasmussen, \emph{{Gaussian Processes for Machine
  Learning}}.\hskip 1em plus 0.5em minus 0.4em\relax MIT Press, 2006.

\bibitem{welch1995introduction}
G.~Welch, G.~Bishop \emph{et~al.}, ``An introduction to the kalman filter,''
  1995.

\bibitem{lampinen2001bayesian}
J.~Lampinen and A.~Vehtari, ``Bayesian approach for neural networks—review
  and case studies,'' \emph{Neural networks}, vol.~14, no.~3, pp. 257--274,
  2001.

\bibitem{bishop2006pattern}
C.~M. Bishop, \emph{Pattern recognition}.\hskip 1em plus 0.5em minus
  0.4em\relax Springer, 2006.

\bibitem{choi2007posterior}
T.~Choi and M.~J. Schervish, ``On posterior consistency in nonparametric
  regression problems,'' \emph{Journal of Multivariate Analysis}, vol.~98,
  no.~10, pp. 1969--1987, Jan. 2007.

\bibitem{box2011bayesian}
G.~E. Box and G.~C. Tiao, \emph{Bayesian inference in statistical
  analysis}.\hskip 1em plus 0.5em minus 0.4em\relax John Wiley \& Sons, 2011.

\bibitem{mcallester1999some}
D.~A. McAllester, ``Some {PAC-Bayesian} theorems,'' \emph{Machine Learning},
  vol.~37, no.~3, pp. 355--363, 1999.

\bibitem{garnett2023bayesian}
R.~Garnett, \emph{Bayesian optimization}.\hskip 1em plus 0.5em minus
  0.4em\relax Cambridge University Press, 2023.

\bibitem{lederer2022cooperative}
A.~Lederer, Z.~Yang, J.~Jiao, and S.~Hirche, ``Cooperative control of uncertain
  multi-agent systems via distributed gaussian processes,'' \emph{IEEE
  Transactions on Automatic Control}, 2022.

\bibitem{liu2018gaussian}
M.~Liu, G.~Chowdhary, B.~C. Da~Silva, S.-Y. Liu, and J.~P. How, ``Gaussian
  processes for learning and control: A tutorial with examples,'' \emph{IEEE
  Control Systems Magazine}, vol.~38, no.~5, pp. 53--86, 2018.

\bibitem{virani2018imitation}
N.~Virani, D.~K. Jha, Z.~Yuan, I.~Shekhawat, and A.~Ray, ``Imitation of
  demonstrations using bayesian filtering with nonparametric data-driven
  models,'' \emph{Journal of Dynamic Systems, Measurement, and Control}, vol.
  140, no.~3, 2018.

\bibitem{mukadam2016gaussian}
M.~Mukadam, X.~Yan, and B.~Boots, ``Gaussian process motion planning,'' in
  \emph{2016 IEEE international conference on robotics and automation
  (ICRA)}.\hskip 1em plus 0.5em minus 0.4em\relax IEEE, 2016, pp. 9--15.

\bibitem{yuan2022dslap}
Z.~Yuan and M.~Zhu, ``{dSLAP}: Distributed safe learning and planning for
  multi-robot systems,'' in \emph{Proc. IEEE Conf. Decision and Control (CDC)},
  2022, pp. 5864--5869.

\bibitem{chiuso2012bayesian}
A.~Chiuso and G.~Pillonetto, ``A bayesian approach to sparse dynamic network
  identification,'' \emph{Automatica}, vol.~48, no.~8, pp. 1553--1565, 2012.

\bibitem{srinivas2012information}
N.~Srinivas, A.~Krause, S.~M. Kakade, and M.~W. Seeger, ``Information-theoretic
  regret bounds for {Gaussian} process optimization in the bandit setting,''
  \emph{IEEE Trans. Information Theory}, vol.~58, no.~5, pp. 3250--3265, Jan.
  2012.

\bibitem{fisac2019general}
J.~F. Fisac, A.~K. Akametalu, M.~N. Zeilinger, S.~Kaynama, J.~Gillula, and
  C.~J. Tomlin, ``A general safety framework for learning-based control in
  uncertain robotic systems,'' \emph{IEEE Trans. Automatic Control}, vol.~64,
  no.~7, pp. 2737--2752, 2019.

\bibitem{yuan2024lightweight}
Z.~Yuan and M.~Zhu, ``Lightweight distributed gaussian process regression for
  online machine learning,'' \emph{IEEE Transactions on Automatic Control},
  2024. To appear.

\bibitem{wilson2016deep}
A.~G. Wilson, Z.~Hu, R.~Salakhutdinov, and E.~P. Xing, ``Deep kernel
  learning,'' in \emph{Artificial intelligence and statistics}.\hskip 1em plus
  0.5em minus 0.4em\relax PMLR, 2016, pp. 370--378.

\bibitem{shen2019random}
Y.~Shen, T.~Chen, and G.~B. Giannakis, ``Random feature-based online
  multi-kernel learning in environments with unknown dynamics,'' \emph{The
  Journal of Machine Learning Research}, vol.~20, no.~1, pp. 773--808, 2019.

\bibitem{vapnik2013nature}
V.~Vapnik, \emph{The Nature of Statistical Learning Theory}.\hskip 1em plus
  0.5em minus 0.4em\relax Springer science \& business media, 2013.

\bibitem{zhou2020general}
Z.~Zhou, O.~S. Oguz, M.~Leibold, and M.~Buss, ``A general framework to increase
  safety of learning algorithms for dynamical systems based on region of
  attraction estimation,'' \emph{IEEE Trans. Robotics}, 2020.

\bibitem{umlauft2020smart}
J.~Umlauft, T.~Beckers, A.~Capone, A.~Lederer, and S.~Hirche, ``Smart
  forgetting for safe online learning with gaussian processes,'' in
  \emph{Learning for dynamics and control}, 2020, pp. 160--169.

\bibitem{berkenkamp2023bayesian}
F.~Berkenkamp, A.~Krause, and A.~P. Schoellig, ``Bayesian optimization with
  safety constraints: safe and automatic parameter tuning in robotics,''
  \emph{Machine Learning}, vol. 112, no.~10, pp. 3713--3747, 2023.

\bibitem{sui2018stagewise}
Y.~Sui, V.~Zhuang, J.~Burdick, and Y.~Yue, ``Stagewise safe bayesian
  optimization with gaussian processes,'' in \emph{Proc. Int. Conf. Machine
  Learning (ICML)}, 2018, pp. 4781--4789.

\bibitem{finn2017model}
C.~Finn, P.~Abbeel, and S.~Levine, ``Model-agnostic meta-learning for fast
  adaptation of deep networks,'' in \emph{International conference on machine
  learning}.\hskip 1em plus 0.5em minus 0.4em\relax PMLR, 2017, pp. 1126--1135.

\bibitem{rajeswaran2019meta}
A.~Rajeswaran, C.~Finn, S.~M. Kakade, and S.~Levine, ``Meta-learning with
  implicit gradients,'' \emph{Proc. Advances in Neural Information Processing
  Systems (NeurIPS)}, vol.~32, 2019.

\bibitem{khodak2019adaptive}
M.~Khodak, M.-F.~F. Balcan, and A.~S. Talwalkar, ``Adaptive gradient-based
  meta-learning methods,'' \emph{Proc. Advances in Neural Information
  Processing Systems (NeurIPS)}, vol.~32, 2019.

\bibitem{lee2021meta}
J.~Lee, J.~Tack, N.~Lee, and J.~Shin, ``Meta-learning sparse implicit neural
  representations,'' \emph{Advances in Neural Information Processing Systems},
  vol.~34, pp. 11\,769--11\,780, 2021.

\bibitem{xu2024online}
S.~Xu and M.~Zhu, ``Online constrained meta-learning: Provable guarantees for
  generalization,'' \emph{Advances in Neural Information Processing Systems},
  vol.~36, 2024.

\bibitem{fallah2021generalization}
A.~Fallah, A.~Mokhtari, and A.~Ozdaglar, ``Generalization of model-agnostic
  meta-learning algorithms: Recurring and unseen tasks,'' \emph{Proc. Advances
  in Neural Information Processing Systems (NeurIPS)}, vol.~34, pp. 5469--5480,
  2021.

\bibitem{yoon2018bayesian}
J.~Yoon, T.~Kim, O.~Dia, S.~Kim, Y.~Bengio, and S.~Ahn, ``Bayesian
  model-agnostic meta-learning,'' \emph{Proc. Advances in Neural Information
  Processing Systems (NeurIPS)}, vol.~31, 2018.

\bibitem{zhang2021shallow}
X.~Zhang, D.~Meng, H.~Gouk, and T.~M. Hospedales, ``Shallow bayesian meta
  learning for real-world few-shot recognition,'' in \emph{Proceedings of the
  IEEE/CVF International Conference on Computer Vision}, 2021, pp. 651--660.

\bibitem{lew2022safe}
T.~Lew, A.~Sharma, J.~Harrison, A.~Bylard, and M.~Pavone, ``Safe active
  dynamics learning and control: A sequential exploration--exploitation
  framework,'' \emph{IEEE Trans. Robotics}, vol.~38, no.~5, pp. 2888--2907,
  2022.

\bibitem{harrison2020meta}
J.~Harrison, A.~Sharma, and M.~Pavone, ``Meta-learning priors for efficient
  online bayesian regression,'' in \emph{Algorithmic Foundations of Robotics
  XIII: Proceedings of the 13th Workshop on the Algorithmic Foundations of
  Robotics 13}, 2020, pp. 318--337.

\bibitem{rothfuss2021pacoh}
J.~Rothfuss, V.~Fortuin, M.~Josifoski, and A.~Krause, ``Pacoh: Bayes-optimal
  meta-learning with pac-guarantees,'' in \emph{Proc. Int. Conf. Machine
  Learning (ICML)}, 2021, pp. 9116--9126.

\bibitem{patacchiola2020bayesian}
M.~Patacchiola, J.~Turner, E.~J. Crowley, M.~O'Boyle, and A.~J. Storkey,
  ``Bayesian meta-learning for the few-shot setting via deep kernels,''
  \emph{Advances in Neural Information Processing Systems}, vol.~33, pp.
  16\,108--16\,118, 2020.

\bibitem{nabi2022bayesian}
S.~Nabi, H.~Nassif, J.~Hong, H.~Mamani, and G.~Imbens, ``Bayesian meta-prior
  learning using empirical bayes,'' \emph{Management Science}, vol.~68, no.~3,
  pp. 1737--1755, 2022.

\bibitem{brunton2016discovering}
S.~L. Brunton, J.~L. Proctor, and J.~N. Kutz, ``Discovering governing equations
  from data by sparse identification of nonlinear dynamical systems,''
  \emph{Proceedings of the national academy of sciences}, vol. 113, no.~15, pp.
  3932--3937, 2016.

\bibitem{dhiman2021control}
V.~Dhiman, M.~J. Khojasteh, M.~Franceschetti, and N.~Atanasov, ``Control
  barriers in bayesian learning of system dynamics,'' \emph{IEEE Trans.
  Automatic Control}, 2021.

\bibitem{wang2018safe}
L.~Wang, E.~A. Theodorou, and M.~Egerstedt, ``Safe learning of quadrotor
  dynamics using barrier certificates,'' in \emph{Proc. Int. Conf. Robotics and
  Automation (ICRA)}, 2018, pp. 2460--2465.

\bibitem{vershynin2018high}
R.~Vershynin, \emph{High-dimensional probability: An introduction with
  applications in data science}.\hskip 1em plus 0.5em minus 0.4em\relax
  Cambridge university press, 2018, vol.~47.

\end{thebibliography}

\end{document}